\documentclass[10pt,twocolumn,letterpaper]{article}

\usepackage{cvpr}
\usepackage{times}
\usepackage{epsfig}
\usepackage{graphicx}
\usepackage{amsmath}
\usepackage{amssymb}
\usepackage{multirow}
\usepackage{tabularx,booktabs}
\usepackage{array,longtable}
\usepackage[normalem]{ulem}
\usepackage{nopageno}

\usepackage[breaklinks=true,bookmarks=false]{hyperref}

\cvprfinalcopy 

\newcommand\redsout{\bgroup\markoverwith{\textcolor{blue}{\rule[0.5ex]{2pt}{0.4pt}}}\ULon}
\newcommand{\rev}[2]{\redsout{}\textcolor{black}{#2}}
\newcommand\tab[1][1cm]{\hspace*{#1}}

\newcommand{\tabincell}[2]{\begin{tabular}{@{}#1@{}}#2\end{tabular}}

\def\cvprPaperID{1082} 

\ifcvprfinal\fi
\begin{document}

\title{Attention-Aware Compositional Network for Person Re-identification}

\author{Jing Xu\textsuperscript{1} \tab Rui Zhao\textsuperscript{1,2}\thanks{Rui Zhao is the corresponding author.} \tab Feng Zhu\textsuperscript{1} \tab Huaming Wang\textsuperscript{1} \tab Wanli Ouyang\textsuperscript{3}\rev{\thanks{Wanli Ouyang is supported by SenseTime Group Limited.}}{} \\
\textsuperscript{1}SenseNets Technology Limited \tab \textsuperscript{2}SenseTime Group Limited \tab  \textsuperscript{3}The University of Sydney \\
{\tt\small eudoraxj@gmail.com, zhaorui@sensetime.com, zhufengx@mail.ustc.edu.cn}  \\
{\tt\small huamingwang2014@gmail.com, wanli.ouyang@sydney.edu.au}
}

\maketitle

\begin{abstract}
\rev{Person}{Person} re-identification (ReID) is to identify pedestrians observed from different camera views based on visual appearance. It is a challenging task due to large pose variations, complex background clutters and severe occlusions. Recently, \rev{predicting human body joint locations by pose estimation}{human pose estimation by predicting joint locations} was largely improved in accuracy. It is reasonable to use pose estimation results for handling pose variations and background clutters, and such attempts have obtained great improvement in ReID performance. However, we argue that the pose information \rev{is}{was} not well utilized and \rev{is not}{hasn't yet been} fully exploited for person ReID. 

In this work, we introduce a novel framework called Attention-Aware Compositional Network (AACN) for person \rev{re-identification}{ReID}. \rev{This framework uses}{AACN consists of two main components:} Pose-guided Part Attention \rev{}{(PPA)} and Attention-aware Feature Composition \rev{}{(AFC)}. \rev{Pose-guided Part Attention is applied to mask pedestrian feature maps to remove undesirable background features.}{PPA is learned and applied to mask out undesirable background \rev{clutters from pedestrian feature maps via an attention mechanism}{features in pedestrian feature maps}}. \rev{It is learned and guided by human pose.}{} Furthermore, pose-guided visibility scores are estimated for body parts to deal with part occlusion \rev{through}{in} \rev{Attention-aware Feature Composition}{the proposed AFC module}. Extensive experiments with ablation analysis show the effectiveness of our method, and state-of-the-art results are achieved on several public datasets, including Market-1501, CUHK03, CUHK01, SenseReID, \rev{}{CUHK03-NP and DukeMTMC-reID}. 
\end{abstract}

\section{Introduction}

\begin{figure}[thb]    
\center{\includegraphics[width=0.47\textwidth]{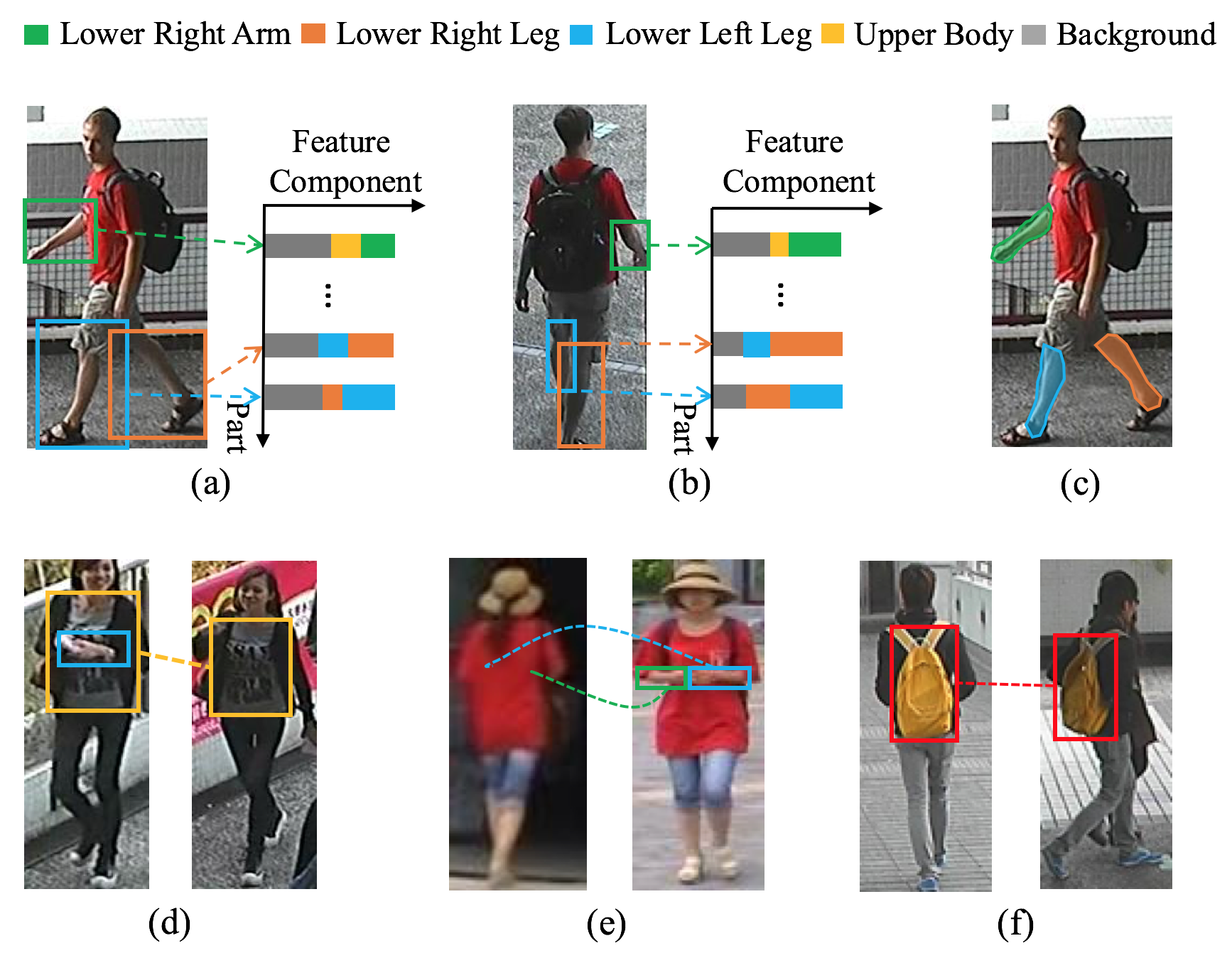}}
\caption{Part alignment challenges in person ReID. (a-c): Describing body parts by bounding boxes may introduce many irrelevant regions from background and other parts. Matching between features extracted from loose boxes in (a) and tight boxes in (b) would deteriorate the matching accuracy. A finer part region representation in (c) would help alleviate this problem. (d-f): The importance of different parts should be adaptively adjusted. Upper body is occluded by forearms in (d), the two forearms are all occluded in (e). Features from occluded part should be eliminated during matching, while salient visual cues like yellow backpack in (f) need to be emphasized.}
\label{fig:introduction} 
\end{figure}


Person re-identification (ReID) targets on identifying the same individual across different camera views. Given an image containing a target person (as query) and a large set of images (gallery set), a ReID system is expected to rank the images from gallery according to visual similarity with the query image. 
It has many important applications in video surveillance by saving large amount of human efforts in exhaustively searching for a target person from large amount of video sequences. For example, finding missing elderly and children, and suspect tracking, \etc. 

Many research works have been proposed to improve the state-of-the-art performance of public ReID benchmarks.
However, identifying the same individual across different camera views is still an unsolved task in intelligent video surveillance. It is difficult in that pedestrian images often suffer from complex background clutters, varying illumination conditions, uncontrollable camera settings, severe occlusions and large pose variations.

\begin{figure}[thb]    
\center{\includegraphics[width=0.47\textwidth]{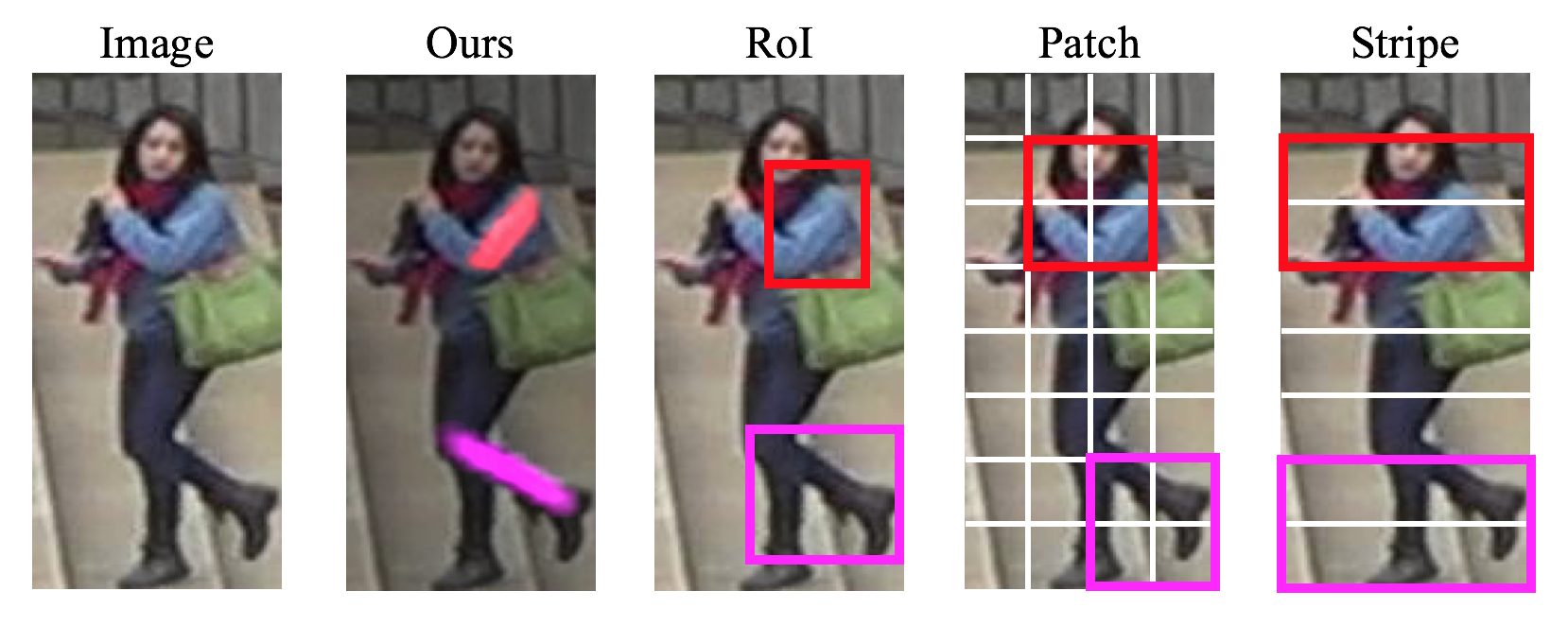}}
\caption{Our Pose-guided Part Attention precisely captures the target parts, excludes background clutter and adjacent part features, while pose-guided rectangular RoIs \cite{zhao2017spindle,su2017pose,zheng2017pose}, patches \cite{zhao2013person,li2014deepreid}, and stripes \cite{shi2015constrained,ahmed2015improved} include extensive noise features. }
\label{fig:align_cmp} 
\end{figure}

Viewpoint changes and pose variations cause uncontrolled misalignment between pedestrian images. As the improvement of human pose estimation \cite{mpiiandriluka20142d,cao2016realtime}, recent works \cite{su2017pose, zhao2017spindle,zheng2017pose} utilized pose estimation results to align body parts for better matching. Although great improvement in performance was obtained, there are still noticeable problems in these methods. These methods deal with misalignment by extracting features from patches, stripes, or pose-guided region of interest (RoI), where rectangular RoIs often introduce noise from adjacent parts or background in feature and lead to inaccurate matching. For example in Fig. \ref{fig:introduction}(a), features of the right leg\rev{(the second bar)}{} is extracted from its bounding box \rev{(the orange box in the image)}{in orange}, which includes extensive noise from left leg and background\rev{}{, as shown in the second bar of the histogram}. Features of the right arm and left leg also consist of their adjacent parts and background. Furthermore, some body parts have large variations in shape and pose, and rectangular RoI would include inconsistent extent of background clutter and adjacent noise. For example, \rev{the orange box in Fig. \ref{fig:introduction}(a) loosely includes the right leg while the orange box in Fig. \ref{fig:introduction})(b) tightly contains the left leg.}{the right leg is loosely included in the Fig. \ref{fig:introduction}(a) and tightly contained in Fig. \ref{fig:introduction}(b)}. Matching between part features from loose box and tight box in two camera views would definitely deteriorate the matching accuracy. \rev{}{To deal with these problems, finer silhouettes contouring body parts like in Fig. \ref{fig:introduction}(c) are needed, so that part features can be extracted more precisely, alleviating the influence from background clutters and adjacent noises. }


In this work, we \rev{introduce}{propose} \rev{a}{to use} Pose-guided Part Attention instead of rectangular RoI. Pose-guided Part Attention is a confidence map \rev{to}{that could} precisely capture the target part, and exclude background clutter and adjacent part features, as shown in Fig. \ref{fig:align_cmp}. Attention-aware part features can be extracted by applying the part attention \rev{}{mask} on feature maps, and feature alignment by part is \rev{easily}{naturally} achieved. We \rev{}{will} show in experiments that \rev{the}{} attention-aware part features are more accurate and robust, and the aligned pedestrian features are more discriminative than \rev{}{those proposed in} conventional methods.

Occlusion is also a common and severe problem in practical ReID scenario. For example in Fig. \ref{fig:introduction}(d-f), body part may be occluded by other body parts, adjacent persons or things like carrying baggage or trolley. Some observations can be concluded: 1) rigid body parts \rev{}{like head-shoulder, upper torso, lower torso} are often partially occluded by adjacent non-rigid parts \rev{}{like upper arms, lower legs, \etc.} 2) non-rigid body parts suffer heavy self-occlusion and are often fully occluded. 3) occlusion by carrying things is not a bad \rev{news}{situation}, which should be considered as a special part to help re-identification. It would be ideal to weaken features for partially occluded rigid part like \rev{}{the upper body in} Fig. \ref{fig:introduction}(d), eliminate features for fully occluded non-rigid part like \rev{}{the forearms in} Fig. \ref{fig:introduction}(e), and retain features for carrying things like \rev{}{the backpack in} Fig. \ref{fig:introduction}(f). Based on above observations on the occlusion problem, we propose a pose-guided visibility score to measure the occlusion extent for each body part, and it provides image-specific part importance score to decide feature importance in matching. Experimental results show its usefulness in handling occlusion cases.

Based on above motivations, a new Attention-Aware Compositional Network for person re-identification is proposed. The contributions of our work can be summarized in several folds: 
\vspace{-6pt}
\begin{itemize}
	\itemsep -4pt
	\item A unified framework named Attention-Aware Compositional Network (AACN) is proposed to deal with misalignment and occlusion problem in person re-identification.
	\item Pose-guided Part Attention is introduced to estimate finer part attention to exclude adjacent noise. It is designed to capture both rigid and non-rigid body parts simultaneously in a unified framework.
	\item Visibility score is introduced to measure the occlusion extent for each body part. It provides image-specific part importance \rev{score to decide feature importance in matching.}{scores for Attention-aware Feature Composition.} 
	\item Extensive experiments demonstrate that our approach achieve superior performance on several public datasets, including CUHK03 \cite{li2014deepreid}, CUHK01 \cite{li2012human}, Market-1501 \cite{zheng2015scalable}, CUHK03-NP \cite{zhong2017re},  DukeMTMC-reID \cite{ristani2016MTMC} and SenseReID \cite{zhao2017spindle}.
\end{itemize}

\section{Related Work}
\subsection{Person Re-identification}
There are two categories of methods addressing the problem of person re-identification, namely feature representation and distance metric learning. The first category mainly includes the traditional feature descriptors \cite{zhao2014learning,zhao2013unsupervised,zhao2013person,liao2015person,chen2016similarity,matsukawa2016hierarchical,shi2015transferring} and deep learning features \cite{xiao2016learning,wang2016joint,wu2016personnet,cheng2016person,li2014deepreid,varior2016gated}. These approaches dedicate to design view-invariant representations for person images. 
The second category \cite{liao2015efficient,cheng2016person,liao2015person,xiong2014person,hermans2017defense,chen2017beyond,li2013locally,ding2015deep,shi2016embedding} mainly targets on learning a robust distance metric to measure the similarity between images. 

Pedestrian alignment, matching two person images with their corresponding parts, is of non-trivial importance. Existing ReID methods mainly focus on extracting two types of features, namely global features extracted from the whole image \cite{chen2017person,xiao2017joint} and region features generated from local patches \cite{zhao2013person,zheng2015partial,li2017learning}. However, these approaches have not taken the accurate alignment of body regions into consideration. Recently, thanks to the great progress of pose estimation methods \cite{cao2016realtime,chu2017multi} and RPN \cite{ren2015faster}, reliable body parts are able to be acquired, which makes it possible to identify individuals via extracted region. For example, Zhao \emph{et al.} \cite{zhao2017spindle} proposed Spindle Net, \rev{that extracted three level parts by body region proposal network, following by feature extraction and fusion stages.}{that extracted and fused three level part features. Parts were extracted by PRN.} Su \emph{et al.} \cite{su2017pose} proposed a Pose-driven Deep Convolutional model (PDC) that utilized Spatial Transformer Network (STN) to localize and crop body regions based on pre-defined centers. Zheng \emph{et al.} \cite{zheng2017pose} introduced to extract Pose Invariant Embedding (PIE) through aligning pedestrians to standard pose. Alignment \rev{is}{was} done by applying affine transformation to pose estimation results. However, these methods are all based on rigid body regions, which cannot accurately localize human body regions. In our model, non-rigid parts are obtained based on the connectivity between human joints. Thus our model is capable of extracting more precise information \rev{from}{for} each body part\rev{. Occlusion is not considered in these works but handled in our work.}{, and handling occlusion issues.}

\subsection{Human Parsing}\label{sec:related_parsing}
Human parsing \cite{gong2017look,chen2014detect,yamaguchi2012parsing,liang2015human,dong2014towards} is related to our work in that \rev{the ideal parsing results can localize body part more accurately than bounding boxes.}{parsing results can accurately localize body part.} For example,
Gong \emph{et al.} \cite{gong2017look} imposed joint structure loss to improve segmentation results.
Dong \emph{et al.} \cite{dong2014towards} explored pose information 
to guide human parsing. However, we choose to generate non-rigid parts based on connectivity of human keypoints rather than human parsing because of the following reason:
Existing human parsing methods mainly focus on particular scenarios, such as fashion pictures, and the parsing models often show weak generalization on surveillance data. Human pose is easier to label than parsing, and it can be better generalized to surveillance scenario owing to large variance of the datasets \cite{mpiiandriluka20142d,cocolin2014microsoft}.

\subsection{Attention based Image Analysis}
Since the attention mechanism is effective in understanding images, it has been widely used in various tasks, including machine translation \cite{bahdanau2014neural}, visual question answering \cite{xu2016ask}, object detection \cite{ba2014multiple}, semantic segmentation \cite{chen2016attention}, pose estimation \cite{chu2017multi} and person re-identification \cite{liu2017end}. Bahdanau \cite{bahdanau2014neural} and Ba \cite{ba2014multiple} adopted recurrent neural networks (RNN) to generate the attention map for an image region at each step, and combined information from different steps overtime to make the final decision. Chen \emph{et al.} \cite{chen2016attention} introduced an attention mechanism that learned to softly weight multi-scale features at each pixel location. Chu \emph{et al.} \cite{chu2017multi} proposed a multi-context attention model for pose estimation. 
Inspired by the methods mentioned above, we propose to learn attention map to capture human body part, and align features across different person images by masking with part attentions. Our attention map is learned guided by pose estimation, and it can contour the shape of part more precisely than rectangular RoI. Furthermore, the intensity of part attention infers the visibility of each part, which helps to deal with part occlusion issues.

\begin{figure*}[thb]        
\center{\includegraphics[width=\textwidth]{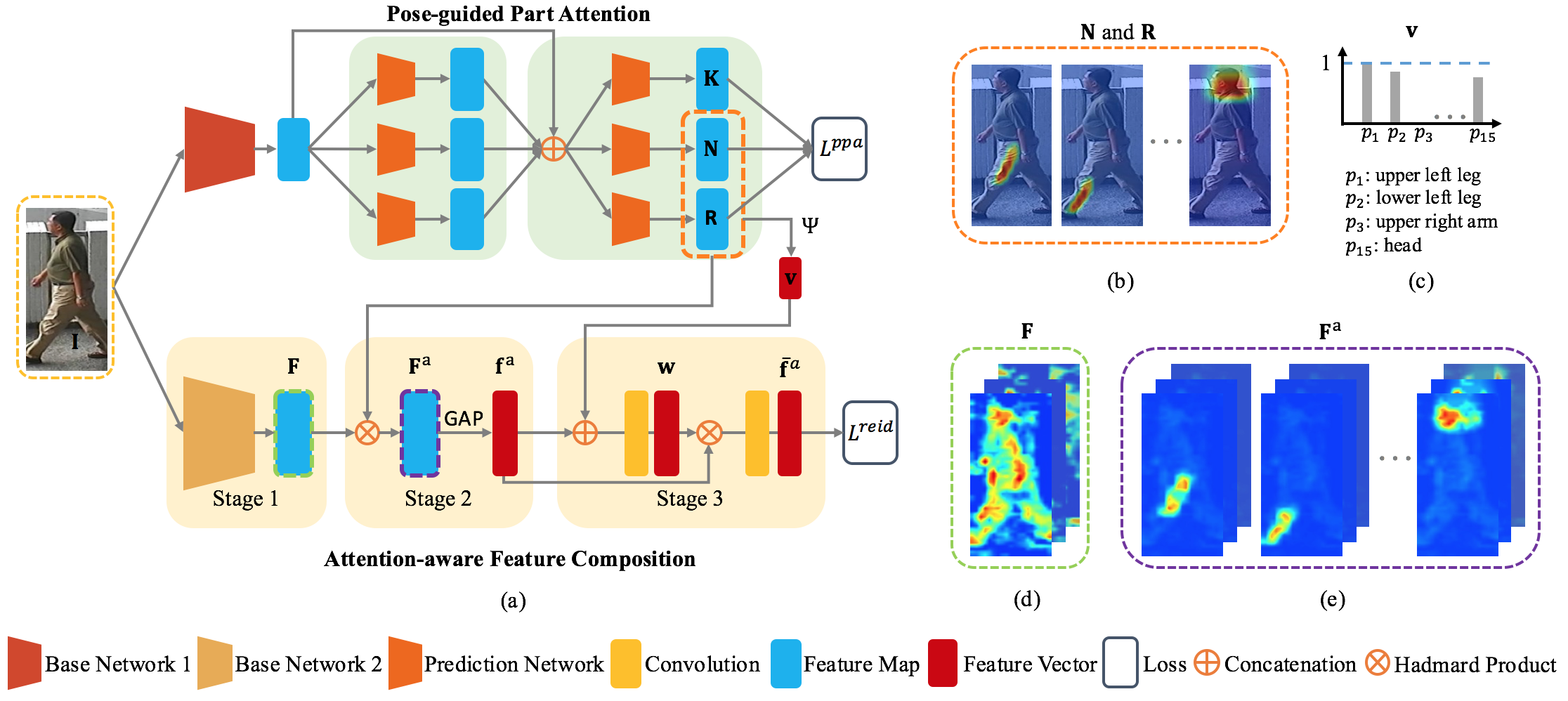}}        
\caption{(a) Attention-Aware Compositional Network (AACN). Our framework consists of two main components: Pose-guided Part Attention (PPA) and Attention-aware Feature Composition (AFC). PPA aims to produce attention maps for locating non-rigid parts $\mathbf{N}$ and rigid parts $\mathbf{R}$. AFC is a 3-stage network that aims to extract robust features for pedestrian images. The first stage generates global context feature maps through a base network. Then, the attention-aware feature maps ${\mathbf{F}^{a}}$ for body parts are extracted in stage 2 with the guidance from part attentions learned in PPA. In stage 3, the part features are further re-weighted by jointly considering part visibility scores $\mathbf{v}$ and feature salience,  resulting in the final compositional weighted feature vector $\bar{\mathbf{f}}^a$. Some visualization are shown in (b) attention maps, (c) visibility scores, (d) global context feature maps, and (e) attention-aware feature maps $\mathbf{F}^{a}$. (Best viewed in color.) }
\label{fig:framework} 
\end{figure*}

\section{Attention-Aware Compositional Network}
\rev{In this section, we illustrate our framework in details. As Fig. \ref{fig:framework} shows, our Attention-Aware Compositional Network (AACN) framework consists of two main components:}{The framework of our Attention-Aware Compositional Network (AACN) is illustrated in Fig. \ref{fig:framework}. AACN consists of two main components:} 1) Pose-guided Part Attention (PPA) and 2) Attention-aware Feature Composition (AFC). Given \rev{a}{one} person image, \rev{our PPA model aims to produce attention map and visibility score for each pre-defined body part}{the proposed PPA module aims to estimate an attention map and a visibility score for each pre-defined body part}. \rev{The AFC model takes in the attention maps and visibility scores, and performs feature alignment and weighted fusion to generate a pedestrian feature}{Then, part feature alignment and weighted fusion are performed in AFC module, given attention maps and visibility scores from PPA}. \rev{The PPA model and the AFC model}{PPA and AFC} are tightly integrated in our framework during both training and testing phases. 


\rev{Our PPA model}{The PPA module} considers two types of pre-defined body parts, \rev{including}{namely,} non-rigid parts and rigid parts. Due to \rev{different properties}{the variations} in appearance, \rev{part }{}attentions \rev{}{of these two types of parts} are estimated separately. The PPA \rev{model takes a three-branch two-stage structure}{module is constructed by a two-stage three-branch neural network}, which predicts confidence maps of keypoints, attention maps of non-rigid parts, and attention maps of rigid parts in \rev{}{the} three branches, respectively. \rev{Furthermore, a}{A} visibility score is \rev{generated}{further estimated} for each part based on \rev{the estimated}{} part attention \rev{, and is integrated in feature composition to deal with part occlusion issue}{maps}. 

\rev{Our AFC model is built based on a baseline network which generates discriminative feature maps, and attention maps are applied to mask the feature maps to produce attention-aware features for each body part. The part features are further combined with visibility scores to generate importance weight for each part, and the attention-aware part features are weightedly fused according to the importance weights. A $1024$-dimensional feature vector is finally produced for the input person image.}{The AFC module applies the estimated part attention maps to mask the global feature map produced by a base network (GoogleNet \cite{szegedy2015going} is used in this work). The resulting attention-aware part features are then weightedly fused with the guidance from part visibility scores. The final $1024$-dimensional feature vector is adopted as the representation of the input person image.}

\subsection{Pose-guided Part Attention}
\label{sec:part_att}
\begin{figure}[t]    
\center{\includegraphics[width=0.48\textwidth]{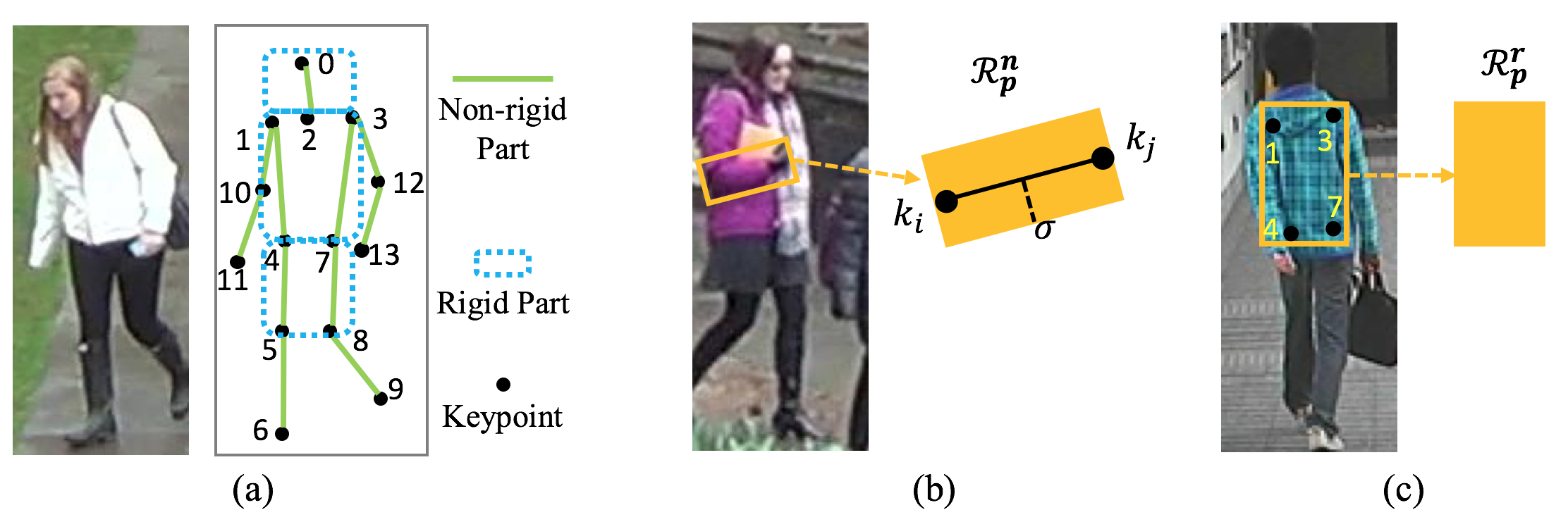}}        
\caption{Illustration of Pose-guided Part Attention. (a) The 14 keypoints, 11 non-rigid parts, and 3 rigid parts defined in our work. (b) The ideal non-rigid part attention $\mathbf{R}_p^n$ for right elbow. (c) The ideal rigid part attention $\mathbf{R}_p^r$ for upper body. }
\label{fig:upa} 
\end{figure}

\rev{In our work, p}{P}art attentions are denoted by normalized part confidence maps, which highlight specific regions of human body in the image. As shown in Fig. \ref{fig:upa}(a), there are two types of human body parts: rigid parts and non-rigid parts. Limb regions including upper arms, lower arms, upper legs, and lower legs are called non-rigid parts because of drastic pose variations they could occur, while trunk parts of human body including head-shoulder, upper torso, and lower torso are considered to be rigid. Attention maps of the two types of \rev{human body }{}parts are simultaneously learned in a unified form through our proposed Pose-guided Part Attention network.

Inspired by the multi-stage CNN \cite{cao2016realtime} for human pose estimation, we utilize a two-stage network to learn part attentions. The first stage individually predicts non-rigid part attentions $\mathbf{N}$, rigid part attentions $\mathbf{R}$, and keypoint confidence maps $\mathbf{K}$ by three independent prediction networks,
\begin{align} 
\label{eq:att_s_1}
\mathbf{N}^1 = \rho^1(\mathbf{F}^{ppa}),~\mathbf{R}^1 = \phi^1(\mathbf{F}^{ppa}),~\mathbf{K}^1 = \psi^1(\mathbf{F}^{ppa}),
\end{align}
where $\mathbf{F}^{ppa}$ is the feature map at the 10-th layer of VGG-19 \cite{vggSimonyan2014Very}. \rev{The task of keypoint estimation is introduced to}{Keypoint estimation is introduced as an auxiliary task to} improve part attention learning in a multi-task learning manner. Then, the second stage refines the attention maps by considering all previous predictions,
\begin{align} 
\label{eq:att_s_2}
\begin{split}
\mathbf{N}^2 &= \rho^2(\mathbf{F}~|~\mathbf{N}^1, \mathbf{R}^1, \mathbf{K}^1), \\
\mathbf{R}^2 &= \phi^2(\mathbf{F}~|~\mathbf{N}^1, \mathbf{R}^1, \mathbf{K}^1), \\
\mathbf{K}^2 &= \psi^2(\mathbf{F}~|~\mathbf{N}^1, \mathbf{R}^1, \mathbf{K}^1).
\end{split}
\end{align}

For network training, supervision is imposed in both stages. The overall objective is
\begin{equation}\label{eq:unifiedloss}
L^{ppa}(\rho, \phi, \psi) = 
\sum_{t=1,2} L^{k}(\mathbf{K}^t) + \mu_1 L^{n}(\mathbf{N}^t) + \mu_2 L^{r}(\mathbf{R}^t\rev{, \mathbf{N}^t}{}),
\end{equation}
where $L^{k}$, $L^{n}$ and $L^{r}$ denote the loss function of keypoint confidence map, non-rigid part attention, and rigid part attention, respectively. $\mu_1$ and $\mu_2$ balance the importance of different losses.

\vspace{1.3mm}
\noindent\textbf{Loss for Keypoint Confidence Map $L^{k}(\mathbf{K})$}.
Following the definition in MPII dataset \cite{mpiiandriluka20142d}, $14$ keypoints (as shown in Fig. \ref{fig:upa}(a)) of human body are utilized to guide the learning of part attentions. The $i$-th channel $\mathbf{K}_i \in \mathbb{R}^{H \times W}$ of keypoint confidence maps $\mathbf{K}\in \mathbb{R}^{H \times W\times C^k}$ predicts the coordinates of the $i$-th keypoint by giving high confidence values to the true location. The difference between confidence maps $\mathbf{K}$ and ground truth maps $\mathbf{K}_i^{*}$ are measured by Mean-Square Error (MSE),
\begin{equation}\label{equ:loss_keypoints}
L^{k}(\mathbf{K}) = \frac{1}{C^k} \sum_{i=1}^{C^k} ||\mathbf{K}_i^* - \mathbf{K}_i||^2,
\end{equation}
where, $\mathbf{K}_i^*$ is generated by applying a Gaussian kernel centered at the true location of the $i$-th keypoint. $C^k = 14$ is the number of keypoints.

\vspace{1.3mm}
\noindent\textbf{Loss for Non-Rigid Part Attention $L^{n}(\mathbf{N})$}.
Non-rigid part attentions aim to highlight the corresponding limb parts. Inspired by the Part Affinity Field (PAF) in \cite{cao2016realtime}, we define the ground truth non-rigid parts as the connection area of two keypoints to approximate the target limb part. As shown in Fig. \ref{fig:upa}(b), the $p$-th non-rigid part is defined as a rectangle area $\mathcal{R}_p^n$ connecting two keypoints with bandwidth $\sigma$, and the ground truth non-rigid part attention is represented as
\begin{align} 
	\mathbf{N}_{p}^{*}(\mathbf{x}) =
		\begin{cases}
		1, & \text{if}\ \mathbf{x}\ \in \mathcal{R}_p^n, \\
		0, & \text{otherwise,}
		\end{cases}
\end{align}
where, $\mathbf{x}$ indicates the location on the attention map. The errors of non-rigid part attention are measured by MSE
\begin{align}\label{equ:nonrigid}
L^{n}(\mathbf{N}) = \frac{1}{C^{n}} \sum_{p=1}^{C^{n}} ||\mathbf{N}^*_p - \mathbf{N}_p||^2,
\end{align}
where, $C^{n} = 11$ is the number of non-rigid parts. $\mathbf{N}_p\in \mathbb{R}^{H \times W}$ is the predicted attention map for the $p$-th part.

\vspace{1.3mm}
\noindent\textbf{Loss for Rigid Part Attention $L^{r}(\mathbf{R}\rev{, \mathbf{N}}{})$}.
Rigid part attention is introduced to capture body parts that take \rev{little non-}{}rigid transformations during changes of view or pose. Three rigid parts are defined in our work, namely head-shoulder, upper torso and lower torso. As shown in Fig. \ref{fig:upa}(c), each rigid part is defined by a \rev{convex polygon which takes human body keypoints as its vertices.}{neat rectangle $\mathcal{R}^r_p$, which tightly contains a set of specified keypoints.} The set of keypoints for each rigid part are selected as $S_1 = \{0,1,3\}$ for head shoulder, $S_2 = \{1,3,4,7\}$ for upper torso, and $S_3 = \{4,5,7,8\}$ for lower torso. \rev{For the $p$-th rigid part attention, the interest region is defined as 
\begin{align}
	\mathcal{R}^r_p = \{\hat{\mathbf{x}}_i ~|~ \hat{\mathbf{x}}_i = s_e\cdot\mathbf{x}_i + (1-s_e)\bar{\mathbf{x}}_i,\nonumber\\
	 \forall \mathbf{x}_i \in Cvx(S_p)\}, \\
	\bar{\mathbf{x}}_i = \frac{1}{|S_p|}\sum_{\mathbf{x}_i \in S_p}\mathbf{x}_{i}, \qquad\qquad\quad
\end{align}
where $|S_p|$ is the number of keypoints in $S_p$, $Cvx(S_p)$ is the convex polygon of the set of keypoints $S_p$, $s_e$ is an expansion factor that is set to $1.2$ in our experiments, and $\bar{\mathbf{x}}_i$ is the center position of all keypoints in $S_p$. }{}
Then the ground truth attention map of rigid part $p$ is defined as 
\begin{equation}\label{equ:rigid}
	\mathbf{R}_p^*(\mathbf{x}) = 
		\begin{cases}
		1, & \text{if}\ \mathbf{x} \in \mathcal{R}^r_p \rev{~\text{and}~ \Sigma_p\mathbf{N}^*_p(
		\mathbf{x}) \leq  \theta }{},\\
		0, & \text{otherwise}.
		\end{cases}
\end{equation}
\rev{where the threshold $\theta$ controls the overlap between rigid part $p$ and all non-rigid parts. We set $\theta=0.1$ in our experiments. }{}The loss for rigid part attention $L^{r}(\mathbf{R})$ is computed by accumulating all part losses,
\begin{align}\label{equ:loss_rigid}
	L^{r}(\mathbf{R}, \mathbf{N}) = \frac{1}{C^r} \sum_{p=1}^{C^r} ||\mathbf{R}^*_p - \hat{\mathbf{R}}_p||^2, \rev{\\
	\hat{\mathbf{R}}_p = \mathbf{R}_p \circ (\Sigma_p \mathbf{N}_p \leq r), \qquad}{}
\end{align}
where $C^r=3$ is the number of rigid parts\rev{, and $\circ$ performs element-wise production on two matrices. $\hat{\mathbf{R}}_p$ is refined prediction of rigid part attention obtained by masking out locations that have non-zero response on non-rigid parts. }{.}

\noindent\textbf{Part Visibility Score}. The intensities in an attention map indicate the visibility of the part at each location. Motivated by this observation, we can define a global visibility score for each part as
\begin{equation}
v_p = \sum_{x,y} |\mathbf{R}_p(x,y)|,~~\text{or}~~v_p = \sum_{x,y} |\mathbf{N}_p(x,y)|,
\end{equation}
where $(x,y)$ indicates the location on the attention map. Global visibility scores help to balance the importance between different body parts for person identification.

\subsection{Attention-aware Feature Composition}
Based on human part attentions introduced in Sec. \ref{sec:part_att}, in this section, we propose an Attention-aware Feature Composition (AFC) which learns to align and re-weight features of body parts. AFC comprises of three main stages, namely Global Context Network (GCN), Attention-Aware Feature Alignment, and Weighted Feature Composition. At the first stage, a pedestrian image is input in GCN to extract global features, which are further fed into stage 2 together with part attentions estimated from the same input image. Stage 2 generates part-attention-aware features and \rev{aligns}{concatenates} them in the same order for all images. The aligned features are then re-weighted by visibility scores in stage 3 to generate the final compositional feature vector. The whole flowchart of AFC is shown in Fig. \ref{fig:framework}. 

\noindent\textbf{Stage 1: Global Context Network (GCN)}.
GCN serves as a base network for global pedestrian feature extraction. Following \cite{geng2016deep}, we build GCN based on the standard GoogleNet \cite{szegedy2015going}. To reduce the computation cost for following stages of AFC, we add one more $256$-channel $3\times3$ convolution layer after the layer ``inception\_5b/output'' of GoogleNet, and then feed the 256-channel feature maps to the next stage. To better fit the aspect ratio of pedestrian images, the input image size is changed from $224\times224$ to $448 \times 192$, and the resulting feature maps at the last convolution layer changes from $7\times 7$ to $14\times 6$ accordingly. 

GCN is independently trained first, and then jointly fine-tuned with following stages of AFC. In independent training, GCN is initialized with the GoogleNet model pre-trained on ImageNet, and the newly added convolution layer is randomly initialized. \rev{\rev{By}{After} globally average pooling the feature maps at the last convolution layer of GCN,}{Subsequent to the global average pooling layer at the end of GCN,} identification loss and verification loss similar to \cite{geng2016deep} are applied to guide the global context learning. In joint fine-tuning, these two losses are retained to preserve high-quality global context features of pedestrian images.

\noindent\textbf{Stage 2: Attention-Aware Feature Alignment}.
Global context features suffer from body part misalignment between pedestrian images. Based on human part attention introduced in Section \ref{sec:part_att}, we propose a simple and effective scheme to achieve attention-aware feature alignment. Specifically, we extract attention-aware feature maps by applying Hadamard Product between global feature maps and each human part attention map, and the resulting feature maps are globally average pooled and concatenated to produce \rev{the}{an} aligned feature vector. Formally, we formulate the attention-aware feature alignment scheme as:
\begin{align}
\mathbf{f}^{a} = Concat(\{\mathbf{f}_p\}_{p=1}^P),~ \mathbf{f}_p = \sigma_{gap}(\mathbf{F}^{a}_p), \\
\mathbf{F}^{a}_p = \mathbf{F} \circ \bar{\mathbf{M}}_p,~ 
\bar{\mathbf{M}}_p = \frac{\mathbf{M}_p}{\text{max}(\mathbf{M}_p)}, \quad~~ 
\end{align}
where, $\mathbf{M}_p \in \{\mathbf{N}_p, \mathbf{R}_p\}$ is the attention map for body parts, $\bar{\mathbf{M}}_p$ is the normalized attention map, $\text{max}(\mathbf{M}_p)$ indicates the maximum value in $\mathbf{M}_p$, $\mathbf{F}$ is the $256$-channel global feature map produced by GCN, $\circ$ denotes the Hadamard Product operator which performs element-wise product on two matrices or tensors, and $\mathbf{F}^{a}_p$ denotes the attention-aware feature map for part $p$. $\sigma_{gap}(\cdot)$ is the global average pooling function, and $Concat(\cdot)$ represents concatenation operation on part feature vectors. Global feature maps $\mathbf{F}$ are masked with attention maps for $P$ times, in which manner $P$ sets of attention-aware feature maps are produced, \ie $\{\mathbf{F}^{a}_p\}_{p=1}^P$. Each set of attention-aware feature maps $\mathbf{F}^{a}_p$ \rev{extract the features centered at}{possess the feature information of} the corresponding body part while preserving global context information of the image. 

The attention-aware feature maps $\mathbf{F}^{a}_p$ are passed through a global average pooling $\sigma_{gap}(\cdot)$ to generate summarized feature vector $\mathbf{f}_p$ for part $p$. These summarized part features are further concatenated to produce final attention-aware aligned feature vector $\mathbf{f}^{a}$ for the input pedestrian image.

\noindent\textbf{Stage 3: Weighted Feature Composition}.
Since pedestrians vary in pose, suffer from occlusions, and may contain some relatively \rev{prominence}{salient} parts, the importance of each part should be adaptively adjusted \rev{while matching two images}{in matching}. Motivated by these observations, we introduce a weight vector $\mathbf{w}$ to measure the part importance. The weight vector is estimated by jointly considering part visibility and feature salience. As shown in Fig. \ref{fig:framework}, visibility scores and the attention-aware aligned feature vector are concatenated, and fed into a fully connected layer (implemented as $1 \times 1$ convolutions) to generate  $\mathbf{w}$. Then the compositional weighted feature vector is generated as $\bar{\mathbf{f}}^a = Conv(Concat(\{\mathbf{w}_p \cdot \mathbf{f}_p\}_{p=1}^P))$, where $Conv$ indicates \rev{convolutions}{a convolution layer}.

Overall, our framework integrated the PPA and ACF to extract feature for each input person image. \rev{Given a query image in application of person ReID, its feature is matched with feature of each image in gallery set, and the distance score is computed.}{In person ReID applications, given a query image, its feature is matched with that of each image in gallery set to generate distance score.} \rev{Gallery images are sorted in ascending order according to the distance scores.}{Gallery images are sorted according to ascending order of the distance scores.} Then the target person can be found among top-ranked gallery images. 

\subsection{Implementation Details} 
 In AFC, GoogleNet is utilized as base network for global context feature extraction, and two additional ``1x1'' convolution layers are used for part weight estimation and final feature fusion, respectively.
 \rev{Furthermore, }{}AACN is trained progressively. First, PPA and GCN are trained independently. PPA is trained with losses for part attention and pose estimation. GCN is trained with reid loss. Then, by fixing PPA and GCN, the parameters for feature weighting and composition in AFC are trained with reid loss. Finally, all modules are jointly fine-tuned.


\begin{table}[thb]
\small
\begin{center}
  \begin{tabular}{cccc}
\hline
Dataset & \#ID & \#train/test IDs & det./lab. \\
\hline
CHUK03 \cite{li2014deepreid} & 1467 & 1160/100 & det.\&lab. \\
CUHK01 \cite{li2012human} & 971 & 486/485 & lab. \\
Market-1501 \cite{zheng2015scalable} & 1501 & 751/750 & det. \\
CUHK03-NP \cite{zhong2017re} & 1467 & 767/700 & det.\&lab. \\
DukeMTMC-reID \cite{ristani2016MTMC} & 1812 & 702/702 & lab. \\
SenseReID \cite{zhao2017spindle} & 1717 & 0/1717 & det. \\
\hline
  \end{tabular}
\end{center}
\vspace{-0.1cm}
\caption{Details of the datasets evaluated in our experiments. Bounding box labels of these datasets can be detected (det.) or manually labeled (lab.).}
\label{table:dataset}
\end{table}


\begin{table}[thb]
\small
\begin{center}
  \begin{tabular}{c|cccc}
     \hline
     CUHK03 (labeled) & R-1 & R-5 & R-10 & R-20\\
     \hline
     NFST \cite{zhang2016learning}      & 62.55 & 90.05 & 94.80 & 98.10 \\ 
     JSTL \cite{xiao2016learning}       & 75.30 & - & - & - \\
     Transfer \cite{geng2016deep}      & 85.40 & - & - & - \\
     SVDNet \cite{sun2017svdnet}        & 81.80 & - & - & - \\ 
     Quadruplet \cite{chen2017beyond}   & 75.53 & 95.15 & 99.16 & - \\ 
     PAR \cite{zhao2017deeply}          & 85.40 & 97.60 & 99.40 & 99.90 \\ 
     \hline
     Spindle \cite{zhao2017spindle}     & 88.50 & 97.80 & 98.60 & 99.20 \\ 
     PDC \cite{su2017pose}              & 88.70 & 98.61 & 99.24 & 99.67 \\ 
     \hline
     AACN (Ours) & \textbf{91.39} & \textbf{98.89}& \textbf{99.48} & \textbf{99.75} \\
     \hline
  \end{tabular}
\end{center}
\vspace{-0.1cm}
\caption{Comparison results on CUHK03 (labeled). }
\label{table:cp_cuhk03_labeled}
\end{table}

\begin{table}[thb]
\small
\begin{center}
  \begin{tabular}{c|cccc}
     \hline
     CUHK03 (detected) & R-1 & R-5 & R-10 & R-20\\
     \hline
     NFST \cite{zhang2016learning}   & 54.70 & 84.75 & 94.80 & 95.20 \\ 
     Transfer \cite{geng2016deep}   & 84.10 & - & - & - \\
     DPFL \cite{chen2017person}      & 82.00 & - & - & - \\
     PAR \cite{zhao2017deeply} & 81.60 & 97.30 & 98.40 & 99.50 \\
     \hline
     PIE \cite{zheng2017pose}  & 67.10 & 92.20 & 96.60 & 98.10 \\
     PDC \cite{su2017pose}     & 78.29 & 94.83 & 97.15 & 98.43 \\
     \hline
	 AACN (ours) & \textbf{89.51} & \textbf{97.68} & \textbf{98.77} & \textbf{99.34}  \\
     \hline
  \end{tabular}
\end{center}
\vspace{-0.1cm}
\caption{Comparison results on CUHK03 (detected).}
\label{table:cp_cuhk03_detected}
\end{table}

\begin{table}[thb]
\small
\begin{center}
  \begin{tabular}{c|cccc}
     \hline
     CUHK01 & R-1 & R-5 & R-10 & R-20\\
     \hline
     NFST \cite{zhang2016learning}     & 69.09 & 86.90 & 91.77 & 95.39 \\ 
     JSTL \cite{xiao2016learning}       & 66.60 & - & - & - \\
     Transfer \cite{geng2016deep}     & 77.00 & - & - & - \\
     Quadruplet \cite{chen2017beyond}  & 62.55 & 83.44 & 89.71 & - \\ 
     PAR \cite{zhao2017deeply}         & 75.00 & 93.50 & 95.70 & 97.70 \\ 
     \hline
     Spindle \cite{zhao2017spindle}    & 79.90 & 94.40 & 97.10 & 98.60 \\ 
     \hline
     AACN (Ours) & \textbf{88.07} & \textbf{96.67} & \textbf{98.16} & \textbf{99.10} \\
     \hline
  \end{tabular}
\end{center}
\vspace{-0.1cm}
\caption{Comparison results on CUHK01. }
\label{table:cp_cuhk01}
\end{table}

\begin{table}[thb]
\small
\begin{center}
  \begin{tabular}{c|cc|cc}
     \hline
     \multirow{2}{*}{Market-1501}
     & \multicolumn{2}{c|}{Single Query} & \multicolumn{2}{c}{Multiple Query} \\
     \cline{2-5} 
      & R-1 & mAP & R-1 & mAP \\
     \hline
     NFST \cite{zhang2016learning}    & 61.02 & 35.68 & 71.56 & 46.03 \\
     PAN \cite{zheng2017pedestrian}   & 82.81 & 63.35 & 88.18 & 71.72 \\
     SVDNet \cite{sun2017svdnet}      & 82.30 & 62.10 & - & - \\
     PAR \cite{zhao2017deeply}           & 81.00 & 63.40 & - & -  \\
     \hline
     Spindle \cite{zhao2017spindle}      & 76.90 & - & - & -  \\
     PIE \cite{zheng2017pose}            & 78.65 & 53.87 & - & -  \\
     PDC \cite{su2017pose}               & 84.14 & 63.41 & - & -  \\
     \hline
     AACN (Ours)  & 85.90 & 66.87 & 89.78 & 75.10 \\
     AACN+R.E.  (Ours)  & \textbf{88.69} & \textbf{82.96} & \textbf{92.16} &  \textbf{87.32} \\
     \hline
  \end{tabular}
\end{center}
\vspace{-0.1cm}
\caption{Comparison results on Market-1501. Rank-1 accuracy (\%) and mAP (\%) are shown. R.E. : re-ranking method from \cite{zhong2017re}.}
\label{table:cp_market}
\end{table}

\begin{table}[thb]
\small
\begin{center}
  \begin{tabular}{c|cc|cc}
     \hline
     \multirow{2}{*}{CUHK03-NP}
     & \multicolumn{2}{c|}{labeled} & \multicolumn{2}{c}{detected} \\
     \cline{2-5}
      & R-1 & mAP & R-1 & mAP \\
     \hline
     PAN \cite{zheng2017pedestrian}      & 36.86 & 35.03 & 36.29 & 34.00 \\
     DPFL \cite{chen2017person}           & 43.00 & 40.50 & 40.70 & 37.00 \\
     SVDNet \cite{sun2017svdnet}          & 40.93 & 37.83 & 41.50 & 37.30 \\
     \hline
     AACN (Ours) & \textbf{81.86} & \textbf{81.61} & \textbf{79.14} & \textbf{78.37}  \\
     \hline
  \end{tabular}
\end{center}
\vspace{-0.1cm}
\caption{Comparison results on CUHK03-NP. }
\label{table:cp_cuhk03_np}
\end{table}

\begin{table}[thb]
\small
\begin{center}
  \begin{tabular}{c|cc}
     \hline
     DukeMTMC-reID & R-1 & mAP \\
     \hline
     OIM \cite{xiao2017joint} & 68.10 & - \\
     LSRO   \cite{zheng2017unlabeled}  & 67.68 & 47.13 \\
     PAN   \cite{zheng2017pedestrian} & 71.59 & 51.51 \\
     SVDNet \cite{sun2017svdnet}       & 76.70 & 56.80 \\
     \hline
     AACN (Ours) & \textbf{76.84} & \textbf{59.25} \\
     \hline
  \end{tabular}
\end{center}
\vspace{-0.1cm}
\caption{Comparison results on DukeMTMC-reID.}
\label{table:cp_duke}
\end{table}

\begin{table}[thb]
\small
\begin{center}
\begin{tabular}{c|cccc}
	\hline
	SenseReID & R-1 & R-5 & R-10 & R-20 \\
	\hline
	JSTL \cite{xiao2016learning} & 23.00 & 34.80 & 40.60 & 46.30 \\
	Spindle \cite{zhao2017spindle} & 34.60 & 52.70 & 59.90 & 66.70 \\
	\hline
	AACN (Ours) & \textbf{41.37} & \textbf{58.65} & \textbf{64.71} & \textbf{72.16} \\
	\hline
\end{tabular}
\end{center}
\vspace{-0.1cm}
\caption{Cross-dataset evaluation on SenseReID.}
\label{table:cross}
\end{table}

\section{Experiments} \label{sec:experiments}
In this section, the performance of AACN is compared with state-of-the-art methods on several public datasets. And then detailed ablation analysis is conducted to validate the effectiveness of AACN components.

\subsection{Datasets and Protocols} \label{sec:dataset}
Our proposed AACN framework is evaluated on several public person ReID datasets, as listed in Table \ref{table:dataset}. For fair comparison, we follow the official evaluation protocols of each dataset. For CUHK03, CUHK01 and SenseReID, Cumulated Matching Characteristics (CMC) at rank-1, rank-5, rank-10 and rank-20 are compared between different approaches. For Market-1501, CUHK03-NP and DukeMTMC-reID, rank-1 identification rate and mean Average Precision (mAP) are reported.

\subsection{Comparisons with State-of-the-Arts} \label{sec:comparison_results}
The proposed AACN is compared with recent approaches with state-of-the-art performance. These approaches are categorized into two sets according to whether human pose information is utilized. One set is pose-irrelevant, which includes 
the null space semi-supervised learning method (NFST) \cite{zhang2016learning},
the domain guided dropout method (JSTL) \cite{xiao2016learning},
the deep transfer learning method (Transfer) \cite{geng2016deep},
the Singular Vector Decomposition method (SVDNet) \cite{sun2017svdnet},
the Online Instance Matching (OIM) method \cite{xiao2017joint},
the quadruplet loss method (Quadruplet) \cite{chen2017beyond},
the multi-scale representation (DPFL) \cite{chen2017person},
the pedestrian alignment network (PAN) \cite{zheng2017pedestrian},
the Part-Aligned Representation (PAR) \cite{zhao2017deeply}.
The other set introduces explicit pose estimation results into ReID, which includes
the Spindle Net (Spindle) \cite{zhao2017spindle}, 
the Pose-driven Deep Convolutional model (PDC) \cite{su2017pose},
and the Pose Invariant Embedding (PIE) \cite{zheng2017pose}.


The experimental results are presented in Table \ref{table:cp_cuhk03_labeled}, \ref{table:cp_cuhk03_detected}, \ref{table:cp_cuhk01}, \ref{table:cp_market}, \ref{table:cp_cuhk03_np} and \ref{table:cp_duke}. It shows that our proposed AACN outperforms state-of-the-art approaches on all datasets. Specifically, when compared with the second best approach on each dataset, our AACN achieves 2.69\%, 5.41\%, 8.17\%, 4.55\% and 40.93\% rank-1 accuracy improvement on CUHK03 (labeled), CUHK03 (detected), CUHK01, Market-1501 and CUHK03-NP (labeled), respectively. Though our AACN is very close to SVDNet \cite{sun2017svdnet} in rank-1 accuracy on DukeMTMC-reID dataset, the improvement in mAP metric (+2.45\%) is still significant.

We also evaluate the generalization ability of our AACN on SenseReID dataset \cite{zhao2017spindle}. Following Spindle \cite{zhao2017spindle}, we merge the training set of Market-1501 \cite{zheng2015scalable}, CUHK01 \cite{li2012human}, CUHK02 \cite{li2013locally}, CUHK03 \cite{li2014deepreid}, PSDB \cite{xiao2016learning}, Shinpuhkan \cite{Shinpuhkan}, PRID \cite{PRID}, VIPeR \cite{gray2007evaluating}, 3DPeS \cite{baltieri20113dpes} and i-LIDS \cite{ilids} for training, and \rev{directly testing}{then test} on SenseReID. As shown in Table \ref{table:cross}, AACN achieves 41.37\% accuracy at rank-1, significantly outperforms Spindle which has an accuracy of 34.60\%.

\begin{table*}[hbtp]
\small
	\begin{center}
		\begin{tabularx}{\textwidth}{c|c|c|c|c|p{0.95cm}<{\centering}|p{0.95cm}<{\centering}|p{0.95cm}<{\centering}|p{0.95cm}<{\centering}|p{0.95cm}<{\centering}|p{0.95cm}<{\centering}}
			\cline{1-11}
            	\multirow{2}{*}{Method} & \multirow{2}{*}{\tabincell{c}{Base model}} & \multirow{2}{*}{\tabincell{c}{\# ince-\\ption}} & \multirow{2}{*}{\tabincell{c}{\# param\\(base)}} & \multirow{2}{*}{\tabincell{c}{\# param\\(overall)}} & \multicolumn{2}{c|}{CUHK03(labeled)} & \multicolumn{2}{c|}{CUHK03(detected)}& \multicolumn{2}{c}{Market-1501(SQ)} \\
			\cline{6-11} 
			& & & & & base & overall & base & overall & base & overall \\
			\cline{1-11}
			\multirow{2}{*}{PIE \cite{zheng2017pose}} 
			& AlexNet & - & 57M & 114M & - & - & 58.80  & 62.60  & 55.49  & 65.68 \\
            \cline{2-11}
			& ResNet-50 & - & 23M & 46M & - & - & 54.80  & 61.50  & 73.02  & 78.65 \\
			\cline{1-11}
			PDC \cite{su2017pose} & GoogleNet-PDC & 10 & 10M & 14M & 79.83  & 88.70  & 71.89  & 78.29  & 76.22  & 84.14 \\
            \cline{1-11}
			Spindle \cite{zhao2017spindle} & GoogleNet-Spindle & 6 & 6M & 44M & - & 88.50  & - & - & 72.10  & 76.90 \\
			\cline{1-11}
			\multirow{2}{*}{AACN (Ours)} & GoogleNet-Spindle & 6 & 6M & 8M & 84.01 & 89.16 & 81.70 & 86.65 & 71.41 & 81.95 \\
            \cline{2-11}
			& GoogleNet & 9 & 6M & 8M & 86.11  & 91.39  & 83.78  & 89.51  & 79.63  & 85.90 \\
			\cline{1-11}
		\end{tabularx}
	\end{center}
    \vspace{-0.1cm}
	\caption{Comparison with human pose based approaches. Rank-1 accuracy (\%) is reported. }
	\label{table:basenet}
\end{table*}

\begin{table}[thb]
\small
\begin{center}
\begin{tabular}{c|ccccc}
	\hline
	Part & Head & L.Arm & R.Arm & L.Leg & R.Leg \\
	\hline
	RoI\cite{zhao2017spindle} & \textbf{26.53} & 13.30 & 13.25 & 13.89 & 14.08 \\
	PPA & 23.51 & \textbf{25.19} & \textbf{23.12} & \textbf{16.63} & \textbf{16.21} \\
	\hline
\end{tabular}
\end{center}
\vspace{-0.1cm}
\caption{Part localization accuracy. Part IoUs are given.}
\label{table:loc}
\end{table}

\begin{table}[thb]
\small
\begin{center}
\begin{tabular}{c|cc|cc}
	\hline
	\multirow{2}{*}{CUHK03}
	& \multicolumn{2}{c|}{labeled} & \multicolumn{2}{c}{detected} \\
	\cline{2-5}
	 & R-1 & R-5 & R-1 & R-5 \\
	\hline
	AFC+RoI\cite{zhao2017spindle} & 89.88 & 97.97 & 86.44 & 97.33 \\
	AFC+Parsing\cite{gong2017look} & 85.49 & 97.38 & 82.92 & 95.66 \\
	\hline
	AFC+PPA & \textbf{90.58} & \textbf{98.65} & \textbf{87.98} & \textbf{97.64} \\
	\hline
\end{tabular}
\end{center}
\vspace{-0.1cm}
\caption{Comparison of part localization methods for ReID.}
\label{table:ppa_select}
\end{table}

\begin{table}[thb]
\small
\begin{center}
\begin{tabular}{c|cc|cc}
	\hline
	\multirow{2}{*}{CUHK03}
	& \multicolumn{2}{c|}{labeled} & \multicolumn{2}{c}{detected} \\
	\cline{2-5}
	 & R-1 & R-5 & R-1 & R-5 \\
	\hline
	GCN & 86.11 & 98.18 & 83.78 & 96.86 \\
	\rev{PPA+AFC\_rigid}{AFC\_rigid} & 87.35 & 97.98 & 84.88 & 96.70 \\
	\rev{PPA+AFC\_non-rigid}{AFC\_non-rigid} & 89.89 & 98.64  & 86.87 & 97.19 \\
	\rev{PPA+AFC\_all}{AFC\_PPA} & \textbf{90.58} & \textbf{98.65} & \textbf{87.98} & \textbf{97.64} \\
	\hline
\end{tabular}
\end{center}
\vspace{-0.1cm}
\caption{Effectiveness of Attention-aware Feature Composition. ``GCN'' uses global features for person ReID. ``AFC\_rigid'' only extracts features from rigid parts.}
\label{table:ppa_component}
\end{table}

\begin{table}[thb]
\small
\begin{center}
\begin{tabular}{c|cc|cc}
  \hline
   \multirow{2}{*}{\tabincell{c}{Rank-1\\accuracy (\%)}} & \multicolumn{2}{c|}{CUHK03} & \multicolumn{2}{c}{Market-1501} \\
	\cline{2-5}
   & labeled  & detected & SQ & MQ \\
  \hline 
  AACN-w/o-v & 90.58 & 87.98 & 86.58  & 90.29  \\
  AACN-v & \textbf{91.39} & \textbf{89.51} & \textbf{88.69}  & \textbf{92.16}  \\  
  \hline
\end{tabular}
\end{center}
\vspace{-0.1cm}
\caption{Effectiveness of visibility score.}
\label{table:ab_ppa_visi}
\end{table}

\subsection{Ablation Analysis} 

\noindent\textbf{Base Network.} The performance of ReID approaches\rev{utilizing human pose cues}{} is influenced by base network structures, and different approaches may choose different backbones. As listed in Table \ref{table:basenet}, our approach is comparable to Spindle \cite{zhao2017spindle} and PDC \cite{su2017pose} in base network size, but much smaller in overall model size. To better compare with previous works on exploiting pose information, we also experiment by replacing base network of our AACN with the one used by Spindle. It shows that our AACN still outperforms Spindle under the same base network structure.

\noindent\textbf{Pose-guided Part Attention.}
The localization accuracy of\rev{ our proposed}{} PPA is compared with rectangular RoI \cite{zhao2017spindle} method on PASCAL-Person-Part dataset \cite{chen2014detect}. Since some body parts are not available on this dataset, we choose head, left arm (L.Arm), right arm (R.Arm), left leg (L.Leg), right leg (R.Leg) for comparison. Localization accuracy is measured as the Intersection-over-Union (IOU) between predictions and ground truth parsing labels. Both methods are trained on the same MPII dataset. As shown in Table \ref{table:loc}, our proposed PPA is more accurate than RoI \rev{method }{}in \rev{locating body parts}{part localization}.

Furthermore, we evaluate the performance of different \rev{body}{part} localization methods on the person ReID task. Specifically, RoI \cite{zhao2017spindle} and parsing results from Parsing \cite{gong2017look} are compared by replacing the PPA module in our AACN framework. The results are shown in Table \ref{table:ppa_select}. On CUHK03 (detected) set, ``AFC+RoI'' is 1.54\% lower than ``AFC+PPA'' in rank-1 accuracy since it includes noise from adjacent areas, and ``AFC+Parsing'' is 5.06\% lower due to domain difference.

\begin{figure}[t]    
\center{\includegraphics[width=0.49\textwidth]{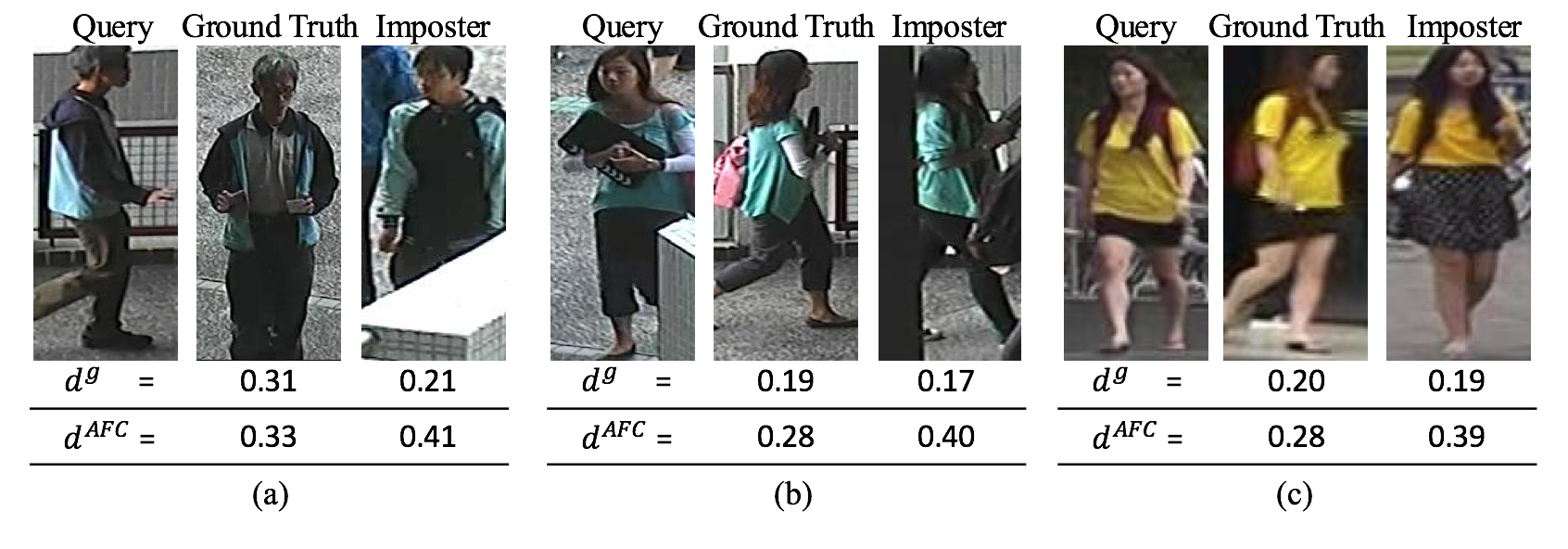}}
\caption{Comparison between global features from GCN and aligned part features from AFC. $d^g$ is the distance computed on global features, and $d^{AFC}$ is the distance computed on the compositional features produced by AFC on Pose-guided Part Attentions. The query images are more similar with the imposters in the global context feature space, but AFC effectively distinguishes them by (a) hair color, (b) upper arm, (c) shorts.} 
\label{fig:ranking}
\end{figure}

\noindent\textbf{Attention-aware Feature Composition.} 
\rev{Attention-aware Feature Composition}{AFC} is a key module in our proposed AACN framework. As shown in Table \ref{table:ppa_component}, ``GCN'' extracts features globally over the image, and achieves 86.11\% and 83.78\% rank-1 accuracy on CUHK03 (labeled) and CUHK03 (detected), respectively. When using AFC, our proposed \rev{``PPA+AFC\_all''}{``AFC\_PPA''} improves the accuracies to 90.58\% and 87.98\%. Using rigid parts only (\rev{``AACN\_rigid''}{``AFC\_rigid''}) or non-rigid parts only (\rev{``AACN\_non-rigid''}{``AFC\_non-rigid''}) still outperforms ``GCN'', and these two types of body parts are complementary to each other. More qualitative results are shown in Fig. \ref{fig:ranking}. Even though a query image looks similar to \rev{the}{an} imposter globally, body part alignment and local feature aggregation in Attention-aware Feature Composition could effectively distinguish them.

\noindent\textbf{Visibility Score.}
The effectiveness of visibility score is evaluated on CUHK03 and Market-1501 dataset. As shown in Table \ref{table:ab_ppa_visi}, weighting part features with visibility scores significantly increase the rank-1 accuracy \rev{on both datasets.}{by 1.53\% and 2.11\% on CUHK03 (detected) and Market-1501 (SQ).}

\section{Conclusion}
In this paper, we propose an Attention-Aware Compositional Network (AACN) to deal with the misalignment and occlusion problem in person re-identification. AACN is composed of two main components, namely, Pose-guided Part Attention (PPA) and Attention-aware Feature Composition (AFC), where PPA is to estimate finer part attention for preciser feature extraction. Also, visibility score is introduced to measure the occlusion extent, and to guide AFC to learn more robust feature for matching. Extensive experiments with ablation analysis demonstrate that our AACN achieves superior performance than state-of-the-art methods on several public datasets. 



\vspace{0.4cm}
\noindent\textbf{Acknowledgements.} This work is supported by SenseTime Group Limited.

{\small
\bibliographystyle{ieee}
\bibliography{egbib}
}

\end{document}